\documentclass[sigconf,nonacm]{acmart}

\usepackage{booktabs} 
\usepackage{stfloats}
\usepackage{graphicx}
\usepackage{amsmath}
\usepackage{booktabs}
\usepackage{multirow}
\usepackage{makecell}
\usepackage{pifont}

\usepackage[ruled]{algorithm2e} 

\begin{document}
\title{Adaptive Local Basis Functions for Shape Completion}

\author{Hui Ying}
\affiliation{%
 \institution{State Key Lab of CAD\&CG, Zhejiang University}
 \streetaddress{866 Yuhangtang Rd}
 \city{Hangzhou}
 \postcode{310058}
 \country{China}}
\email{huiying@zju.edu.cn}

\author{Tianjia Shao}
\authornote{Corresponding author.}
\affiliation{%
 \institution{State Key Lab of CAD\&CG, Zhejiang University}
 \streetaddress{866 Yuhangtang Rd}
 \city{Hangzhou}
 \postcode{310058}
 \country{China}}
\email{tianjiashao@gmail.com}
 
\author{He Wang}
\affiliation{%
 \institution{University of Leeds}
 \streetaddress{Woodhouse Lane}
 \city{Leeds}
 \postcode{LS2 9JT}
 \country{UK}}
\email{H.E.Wang@leeds.ac.uk}

\author{Yin Yang}
\affiliation{%
 \institution{University of Utah}
 \streetaddress{201 Presidents Cir}
 \city{Salt Lake City}
 \postcode{84112}
 \country{USA}}
\email{yin.yang@utah.edu}

\author{Kun Zhou}
\affiliation{%
 \institution{State Key Lab of CAD\&CG, Zhejiang University}
 \streetaddress{866 Yuhangtang Rd}
 \city{Hangzhou}
 \postcode{310058}
 \country{China}}
\email{kunzhou@acm.org}

\begin{abstract}
 In this paper, we focus on the task of 3D shape completion from partial point clouds using deep implicit functions. Existing methods seek to use voxelized basis functions or the ones from a certain family of functions (e.g., Gaussians), which leads to high computational costs or limited shape expressivity. On the contrary, our method employs adaptive local basis functions, which are learned end-to-end and not restricted in certain forms. Based on those basis functions, a local-to-local shape completion framework is presented. Our algorithm learns sparse parameterization with a small number of basis functions while preserving local geometric details during completion. Quantitative and qualitative experiments demonstrate that our method outperforms the state-of-the-art methods in shape completion, detail preservation, generalization to unseen geometries, and computational cost. Code and data are at \url{https://github.com/yinghdb/Adaptive-Local-Basis-Functions}.
\end{abstract}

%
%



\begin{teaserfigure}
  \includegraphics[width=\textwidth]{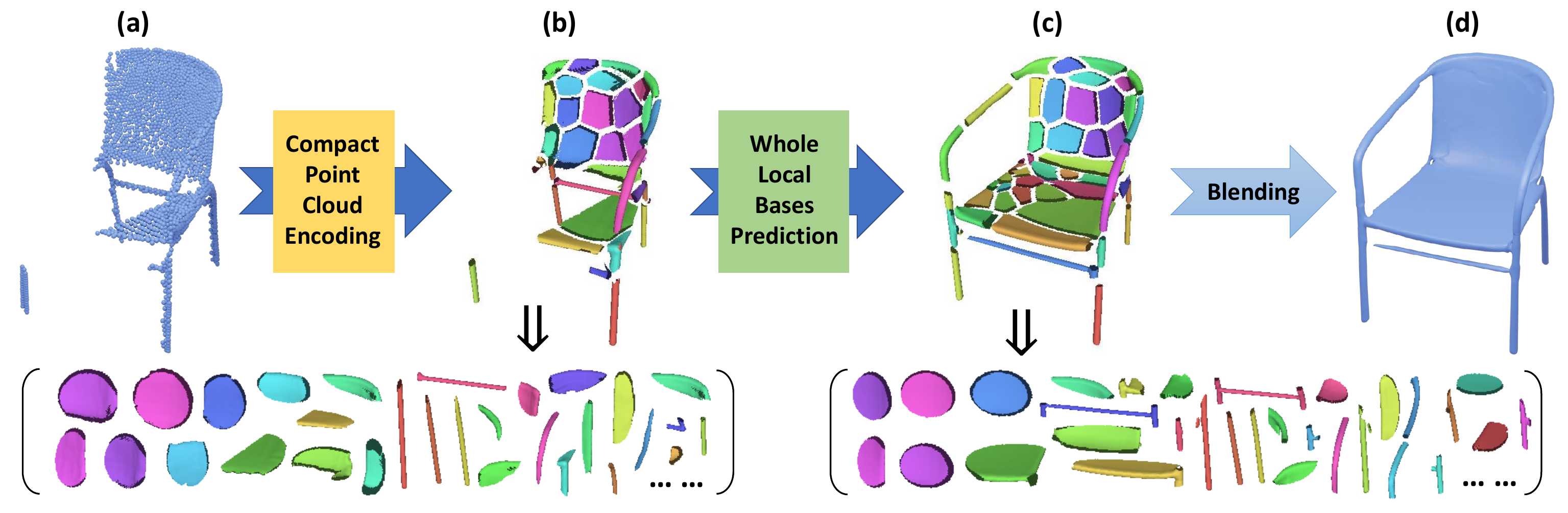}
  \caption{Given an input partial point cloud (a), we first encode the visible observation as a set of local basis functions (32 in this example) (b), and then use them to predict the local basis functions for the overall shape (64 in this example) (c), which are blended to form the final shape (d). We visualize the implicit surfaces inferred from the local basis functions in the bottom of (b) and (c), while (b) and (c) show the positioned local basis functions without the overlapped regions.}
  \label{fig:overall_pipeline}
\end{teaserfigure}

\maketitle

\section{Introduction}
\label{sec:intro}

3D Shape completion from partially scanned point clouds has been widely studied due to its importance to various applications such as automatic driving, augmented reality, and robotics. Naturally, one needs to rely on certain schemes to represent the 3D shapes we want to complete such as point clouds~\cite{mazur2021cloud, wang2022learning, xie2020grnet, liu2020morphing, yuan2018pcn, xiang2021snowflakenet}, deformable meshes~\cite{rock2015completing, litany2018deformable}, and voxels~\cite{choy20163d, hane2017hierarchical, han2017high, dai2017shape}. On the downside, those classic representations also exhibit several intrinsic limitations. For instance, a point cloud often needs extra post-processing; the deformable template mesh may not fit the topology of target object; while processing voxel-based shapes is much more expensive. Deep implicit functions or DIFs have recently attracted more attention, which have been proven to be highly effective for the completion of 3D objects~\cite{park2019deepsdf, mescheder2019occupancy, genova2019learning, genova2020local}. 

Traditionally, an implicit function is regarded as a weighted combination of multiple basis functions~\cite{turk2002modelling, walder2006implicit}. Those basis functions and the associated weights can be computed with respect to an individual geometry. In the context of deep learning, a DIF encodes an input observation using a latent vector $\textbf{z}$ and adopts a network-based embodiment to estimate the function value $f(\textbf{x}, \textbf{z})$ for a given 3D query location $\textbf{x}$. Most existing DIF methods follow this modality but use different choices of implicit functions (i.e., either as a global basis function or multiple local basis functions) and weighting mechanisms, which collectively determine the expressivity of the DIF.

Early DIF methods~\cite{park2019deepsdf, mescheder2019occupancy, genova2019learning} estimate a signed distance or an occupancy function utilizing a single latent code, with a global basis function. This representation is later proven to be limited in describing complex shapes~\cite{genova2020local, chibane2020implicit}. Therefore, researchers switched to localized basis functions for shape completion by dividing the whole shape into multiple regions and region-conditioned latent codes. One line of research is to discretize the space into a regular voxel grid and embed the local latent codes in the voxels~\cite{chibane2020implicit, chen2021multiresolution}. While being able to achieve the state-of-the-art results in shape completion, the required grid resolution leads to a significant growth of computational cost. Alternatively, adaptive parameterization is sought for more compact representations (LDIF)~\cite{genova2020local}, which learns to decompose a shape into a collection of overlapping regions represented by 3D Gaussian basis functions. In each region a latent code is assigned to learn a residual coefficient function for the Gaussian basis function to produce details. Despite the impressive results in shape completion, as the final shape is based on a mixture of refined Gaussians, it inherently has a limited capacity to capture full details, and therefore can still miss geometric details in difficult cases (see~Fig.~\ref{fig:compare} for example).

In this paper, we argue that the specific form of basis functions should be learnable without being restricted to a certain family of functions. Such basis functions can potentially bring multiple benefits. Since the basis functions are learnable and local, they are more likely to capture local fine details due to the data-driven nature. Because the center and the shape of basis functions are learnable, fewer basis functions can achieve equal or better representation for the same geometry compared to analytical functions. For this reason, we aim to learn arbitrarily shaped basis functions so that we can complete 3D shapes with more local details and lower computational costs.

Learning basis functions for shape completion however is a challenging task. It is known that global DIFs can still miss local details~\cite{genova2020local, chibane2020implicit}, and we would like to keep our learned functions local. Enforcing such \emph{locality} is non-trivial during the training process. In addition, unlike learning DIFs from full observations~\cite{chabra2020deep, jiang2020local, yao20213d, li2022learning}, we only have partial observations in shape completion tasks, which impose further difficulties. To this end, our method leverages a progressive, observed-to-unobserved process. We first encode the visible shape as a sequence of local basis functions and then use them to predict the basis functions in the missing region in a sequence-to-sequence manner. This strategy tends to preserve the fine details for the visible area while learning the correlations between the visible and missing parts.

We address the locality problem by learning the function domain of each basis. Following the intuition that points located near a basis should be more likely to be inside its domain, we adopt the Radial Basis Function (RBF) kernels with learnable parameters to parameterize the domain. In our implementation, the RBF-based domains and DIF-based basis functions are learned jointly in an end-to-end manner for a compact shape representation and preserved local details. Based on such shape formulations, we build the shape completion pipeline in two main steps. As shown in Fig.~\ref{fig:overall_pipeline}, the first step is to map the partial input points into a collection of local bases which encode the visible shape compactly with details. In the second step, we predict the local bases for the missing areas and refine the local bases of the visible areas by adopting Transformer encoders~\cite{vaswani2017attention}. The self-attention mechanism of Transformers mimics the pairwise interaction between local bases, thus enabling the accurate sequence-to-sequence translation among visible local bases and missing local bases. 

To summarize, our main contributions include the following aspects. First, we propose DIF-based local basis functions for effective and efficient shape representation, which can capture fine details with a small number of local basis functions. Second, we introduce a local-to-local shape completion pipeline, which is both efficient and geometry-rich. Experiments demonstrate that our method outperforms previous state-of-the-art methods by a large margin.

\section{Related Work}

\subsection{Deep Implicit Shape Representation}

A large amount of learning-based methods has achieved promising results using implicit shape representation. With the strength of deep learning, neural networks serve as a powerful tool to fit various implicit functions, such as signed/unsigned distance fields~\cite{park2019deepsdf, venkatesh2020dude}, occupancy indicator functions~\cite{chen2019learning, mescheder2019occupancy, peng2020convolutional}, deformation functions~\cite{paschalidou2021neural, hui2022neural, deng2021deformed} or other specifically defined implicit functions~\cite{morreale2021neural, chen20223psdf, aumentado2022representing}. 

Pioneering works such as OccNet~\cite{mescheder2019occupancy}, IM-Net~\cite{chen2019learning}, and DeepSDF~\cite{park2019deepsdf} show that many simple shapes can be represented by a latent code and the corresponding deep implicit function. 
However, such deep representations often fail to capture local geometries for more complex shapes. Recent works overcome the problem by focusing on the localized basis functions. 
Some methods divide the 3D space into voxel grids and assign each voxel with a latent code~\cite{chabra2020deep, jiang2020local}, while some store the latent codes in the grid points~\cite{peng2020convolutional, chibane2020implicit, chen2021multiresolution} (or octree~\cite{takikawa2021neural}) and interpolates them for query points within the voxel. Then the local basis functions are learned separately for each voxel and all local bases are combined for the final shape reconstruction. 
While state-of-the-art results can be achieved, increased resolution yields a significant growth in the number of codes, resulting in high computational costs.

In addition to the grid-based DIFs, some methods~\cite{zhang20223dilg, li2022learning, chen2022latent, xiao2022taylorimnet, yao20213d, tretschk2020patchnets, genova2019learning, genova2020local, hertz2022spaghetti} seek to formulate the local bases with irregular positions.
SIF~\cite{genova2019learning} decomposes a shape into a collection of overlapping regions represented by 3D Gaussian basis functions, and LDIF~\cite{genova2020local} learns adaptive weights with DIFs for further refinement. LGCL~\cite{yao20213d} samples a set of key points and divide the 3D space into local regions based on Euclidean distance, within which it learns a DIF for local shape representation. 
Some other methods~\cite{li2022learning, chen2022latent} store the latent codes in multiple irregularly distributed key points, and the new code is interpolated from these codes. 
The advantage of these methods is that less computation is used to represent a complex shape.
In our method, we follow this strategy but propose a novel formulation such that the DIF-based local bases are learned along with the combining weights so as to capture more details.

\subsection{Shape Completion}

Recently, neural networks have been used to predict the whole shape from partial input with the help of data priors. The shape completion methods can be classified according to the output representations, such as voxels, meshes, point clouds, and deep implicit functions. 
Voxel-based methods~\cite{choy20163d, hane2017hierarchical, han2017high, dai2017shape, sun2022patchrd} can directly generate output data thanks to 3D convolution networks, but the memory and time costs are too high when dealing with high-resolution shape grids. 
And mesh-based methods~\cite{rock2015completing, litany2018deformable} are hard to handle shapes with arbitrary topology. 
Therefore, a mass of methods~\cite{mazur2021cloud, wang2022learning, xie2020grnet, liu2020morphing, yuan2018pcn, xiang2021snowflakenet} focus on performing shape completion in point clouds which do not have those problems. But usually what we want are the mesh outputs rather than the point clouds.
Other popular methods recently for shape completion is using DIF~\cite{park2019deepsdf, mescheder2019occupancy, genova2019learning, genova2020local}. 
For preserving details and using the convenient 3D convolution, most methods~\cite{chibane2020implicit, chen2021multiresolution, mittal2022autosdf, zheng2022sdf, yan2022shapeformer} employ grid-based features to process data and express the implicit function of the output shape. 
ShapeFormer~\cite{yan2022shapeformer}, AutoSDF~\cite{mittal2022autosdf} and SDF-StyleGAN~\cite{zheng2022sdf} propose to model the shape completing as a generative task which aims to generate a series of voxelized latent codes for representing the complete shape. 
However, due to the large amount of latent codes and the use of 3D convolution, these methods suffer the problem of high computational cost. 
SIF~\cite{genova2019learning} and LDIF~\cite{genova2020local} perform 2D convolutions on the input partial depth map(s) to extract the features which encode the whole shape. But their Gaussian-based local bases limit their expression capacity of arbitrary shapes.
Without the above issues, our shape completion method utilizes DIF-based local bases with arbitrary shapes to preserve better details and avoid the use of 3D convolution to consume lower computation.

\subsection{Transformers} 

Transformer~\cite{vaswani2017attention} is a powerful framework for sequence-to-sequence translation tasks, which has been proved useful in natural language processing~\cite{devlin2018bert, radford2019language} and image processing~\cite{parmar2018image, dosovitskiy2020image, carion2020end}. 
Most recently, a number of methods~\cite{yu2021pointr, yan2022shapeformer, mittal2022autosdf} model the shape completion as a sequence-to-sequence task by taking advantage of Transformers. PoinTr~\cite{yu2021pointr}, as a point cloud completing method, uses the Transformer encoder-decoder architecture to predict point proxies for missing parts. ShapeFormer~\cite{yan2022shapeformer} and AutoSDF~\cite{mittal2022autosdf}, as implicit-function based methods, use the Transformer-based autoregressive model to predict the complete shape conditioned on the partial inputs. In our method, we utilize the Transformer encoder to model the dependencies among the visible and missing parts and predict the local bases for complete shape representation.

\section{Implicit Field Formulation}

\label{sec:formulation}

A surface can be described as an SDF and represented implicitly as $\{\textbf{x} | f_{\phi}(\textbf{x}, \textbf{z})=0\}$, where $f_{\phi}(\textbf{x}, \textbf{z})$ can be implemented by a neural network with learnable parameters $\phi$. Unlike previous methods which assume a global latent code $\textbf{z}$~\cite{park2019deepsdf}, we represent the SDF as a weighted sum of multiple local basis functions. 
Each DIF $f_{\phi}(\textbf{x}-\boldsymbol{\mu}_i, \textbf{z}_i)$, or $f_i(\textbf{x})$ for simplicity, is defined with a center position $\boldsymbol{\mu}_i$, and a latent code $\textbf{z}_i$ is used for expressing the local SDF. 
For a given query point $\textbf{x}\in \mathbb{R}^3$, its final signed distance is decided by a linear combination of $N$ DIF-based basis functions with weights $\alpha_i$,

\begin{equation}
\begin{aligned}
   sdf(\textbf{x}) = 
  \sum_{i \in [N]} 
  \alpha_i f_{\phi}(\textbf{x}-\boldsymbol{\mu}_i, \textbf{z}_i), \\
   \alpha_i = 
  \frac{g(\textbf{x}- \boldsymbol{\mu}_i, \textbf{A}_i)}
  {\sum_{j\in [N]} g(\textbf{x} - \boldsymbol{\mu}_j,  \textbf{A}_j)}, \\
   g(\textbf{x}-\boldsymbol{\mu}_i, \textbf{A}_i) = 
   \exp{(-||\textbf{A}_i (\textbf{x}_i - \boldsymbol{\mu}_i)||_2^2)},
  \label{eq:sdf_form_2}
\end{aligned}
\end{equation}
where $g(\textbf{x}- \boldsymbol{\mu}_i, \textbf{A}_i)$, or $g_i(\textbf{x})$ for simplicity, is an RBF function with learnable parameter $\textbf{A}_i$. 
$\textbf{A}_i$ is a linear transform matrix, which is constructed by the product of a scaling matrix $\textbf{S}_i$ and a 3D rotation matrix $\textbf{R}_i$. 
Practically, $\textbf{S}_i$ is mapped from a 3-dimensional vector, and $\textbf{R}_i$ is mapped from a 6-dimensional vector as in~\cite{zhou2019continuity}, in which these vectors are predicted from networks directly. 
Eq.~\ref{eq:sdf_form_2} naturally encourages sparsity through $\alpha_i$, which increases exponentially when $\textbf{x}$ is close to $\boldsymbol{\mu}_i$, but quickly becomes damped when $\textbf{x}$ moves away from $\boldsymbol{\mu}_i$. 

For complex shapes, one may still need to use many basis functions to capture local shape variations, and $\alpha_i$ alone is insufficient to guarantee the sparsity.   
To this end, we require each $f_{\phi}$ only parameterizes a local neighborhood around it. As a result, only a small number of $f_{\phi}$s contribute the actual value of $sdf(\textbf{x})$ for a given $\textbf{x}$. 
In our experiments, we found that only two nearest $f_{\phi}$s to $\textbf{x}$ will give reasonably good results.
Let $p$ and $q$ be the indices of two largest RBFs $g_i(\textbf{x})$, and $sdf(\textbf{x})$ becomes the linear combining the local bases with these two indices (i.e., replacing $[N]$ with $\{p, q\}$ in Eq.~\ref{eq:sdf_form_2}).

\begin{figure*}[t]
  \centering
   \includegraphics[width=1.0\linewidth]{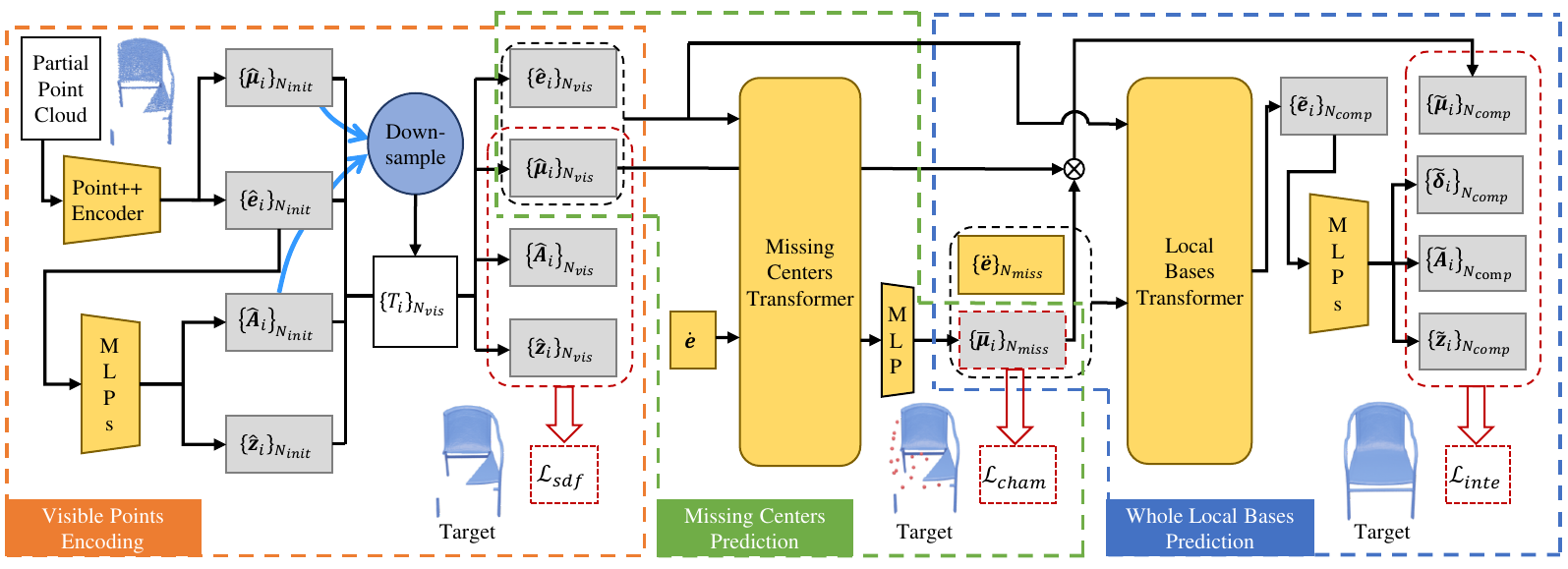}
   \caption{The overall network architecture of shape completion is divided into three parts. `Visible Points Encoding' takes a partial point cloud as input and encodes it into $N_{vis}$ local bases for visible regions, which is supervised by $\mathcal{L}_{sdf}$. `Missing Centers Prediction' predicts $N_{miss}$ centers of local bases for invisible regions based on the encoding and centers of visible local bases, which is supervised by $\mathcal{L}_{cham}$. With the above predictions, `Whole Local Bases Prediction' predicts the final $N_{comp}$ local bases for the complete shape, which is supervised by $\mathcal{L}_{inte}$. Within the figure, yellow blocks stand for learnable networks or parameters, $\dot{e}$ and $\{\ddot{e}\}_{N_{miss}}$ are query embeddings, and $\otimes$ means a concatenate operation. Both Missing Centers Transformer and Local Bases Transformer use the Transformer encoder architecture with multiple self-attention layers. Note that $N_{comp}=N_{vis}+N_{miss}$.} 
   \label{fig:overall_net}
\end{figure*}

\section{Completion Pipeline}

As shown in Fig.~\ref{fig:overall_net}, we first encode the input partial points into a series of local bases as the shape representation for the visible area (see Sec.~\ref{sec:point_cloud_encoding}), and then utilize the power of Transformers~\cite{vaswani2017attention} to generate the whole local bases (see Sec.~\ref{sec:local_shape_prediction}), which can further be optimized for better results (see Sec.~\ref{sec:post_optimization}).

\subsection{Compact Point Cloud Encoding}
\label{sec:point_cloud_encoding}

As shown in the orange dashed box in Fig~\ref{fig:overall_net}, the PointNet++ encoder~\cite{qi2017pointnet++} serves to downsample and encode input points into $N_{init}$ center points with coordinates $\hat{\boldsymbol{\mu}}_i$ and embeddings $\hat{\textbf{e}}_i$. Then Multi-Layer Perceptrons (MLPs) are used to decode the embeddings into latent codes $\hat{\textbf{z}}_i$ and domain parameters $\hat{\textbf{A}}_i$ which together with the centers $\hat{\boldsymbol{\mu}}_i$ form the local bases for the shape representation.

In the PointNet++ encoder, key points are sampled uniformly from the input points. Such sampling is unnecessary. Ideally, regions with complex geometry should be densely sampled while regions with simple geometry should be sparsely sampled, as the domain of the local basis will be smaller for complex geometry and larger for simple geometry. 
Therefore, we propose an adaptive downsampling strategy based on the predicted domains of local bases after the uniform sampling in PointNet++ encoder. The detailed downsampling algorithm is shown in Alg.~\ref{alg:downsample}. 
$g_i(\boldsymbol{\mu}_j)$ can be regarded approximately as the the probability that the center of $j$-th local basis is inside the domain of the $i$-th local basis. 
So the higher $s(j)$ implies the higher probability that the domain of $j$-th local basis can be covered by the other local bases, so we eliminate the key point with the highest $s(j)$ sequentially. Note that after eliminating one key point, all the other $s(j)$ needs to be updated.

\begin{algorithm} 
\caption{Domain-based Downsampling} 
\label{alg:downsample} 
Let $S = \{s(j) = \sum_{i \in [N_{init}], i \neq j} g_i(\boldsymbol{\mu}_j)\}_{j \in [N_{init}]} $ \;
Let $T = [N_{init}]$ be the reserved indices\;
\For{$iter=1$ to $N_{init} - N_{vis}$} {
Find the max $s(k)$ in $S$ \;
Eliminate $k$ from $T$, and $s(k)$ from $S$ \;
Update $s(j) = s(j) - g_k(\boldsymbol{\mu}_j)$ for $s(j) \in T$ \; }
\end{algorithm}

In order to learn the compact encoding for the input partial point cloud, we perform end-to-end training for partial shape representation with the following loss:

\begin{equation}
\begin{aligned}
  \mathcal{L}_{sdf} 
  = \frac{1}{|\mathcal{X}|}\sum_{\textbf{x},\textbf{y} \in \mathcal{X},\mathcal{Y}} 
  \alpha_p |f_{p}(\textbf{x})-\textbf{y}| + 
  \alpha_q |f_{q}(\textbf{x})-\textbf{y}|, \\
  \label{eq:sdf_loss}
\end{aligned}
\end{equation}
where $\mathcal{X}$ and $\mathcal{Y}$ stand for the set of query points and target signed distances, and $\{p, q\}$ are as described in Sec.~\ref{sec:formulation}. For partial point cloud encoding, the query points $\mathcal{X}$ for training are sampled near the input points. 
With the loss function, we want each local basis function $f_{i}$ to accurately learn the local SDF function. And at the same time, 
the learning of $\alpha_i$ and $f_{i}$ is adaptive: if $|f_{i}(\textbf{x})-\textbf{y}|$ is smaller, the corresponding $\alpha_i$ will be learned to be larger to reduce $\mathcal{L}_{sdf}$ since $\alpha_p+\alpha_q=1$. This means that if a position can be more accurately described by a local basis function, it will be more likely in the local domain of the basis.

\subsection{Whole Local Bases Prediction}
\label{sec:local_shape_prediction}

In Sec.~\ref{sec:point_cloud_encoding}, we get $N_{vis}$ centers $\hat{\boldsymbol{\mu}}_i$ and embeddings $\hat{\textbf{e}}_i$ which can be decoded into local bases for visible shape representation. Taking these results as input, we aim to predict more centers and embeddings for complete shape representation in two steps as shown in the green and blue dashed boxes in Fig.~\ref{fig:overall_net}.

\paragraph{Missing Centers Prediction. }
First, the $N_{vis}$ pairs of embeddings $\hat{\textbf{e}}_i$ and centers $\hat{\boldsymbol{\mu}}_i$, as well as one learnable query embedding $\dot{\textbf{e}}$, are input into Missing Centers Transformer, 
and the output is then fed to the MLP to get $N_{miss}$ center points coordinates $\bar{\boldsymbol{\mu}}_i$ for the missing area. 
We supervise the prediction of missing centers with the following chamfer distance loss:

\begin{equation}
\begin{aligned}
  \mathcal{L}_{cham} 
  = \frac{1}{|\mathcal{M}|} \sum_{\boldsymbol{\mu} \in \mathcal{M}} \min_{\boldsymbol{\nu} \in \mathcal{N}} ||\boldsymbol{\mu}-\boldsymbol{\nu}||_2 +\frac{1}{|\mathcal{N}|} \sum_{\boldsymbol{\nu} \in \mathcal{N}} \min_{\boldsymbol{\mu} \in \mathcal{M}} ||\boldsymbol{\nu}-\boldsymbol{\mu}||_2,
  \label{eq:Chamfer_loss}
\end{aligned}
\end{equation}
where $\mathcal{M}$ is the set of $N_{miss}$ predicted missing center point coordinates, and $\mathcal{N}$ is the set of target missing center point coordinates. The target centers are fetched by using Furthest Point Sampling (FPS) in the surface points of the missing region.

\paragraph{Whole Local Bases Prediction. }
Local Bases Transformer takes as input $N_{vis}$ pairs of embeddings $\hat{\textbf{e}}_i$ and centers $\hat{\boldsymbol{\mu}}_i$ for visible region, as well as $N_{miss}$ pairs of query embeddings $\ddot{\textbf{e}}$ and centers $\bar{\boldsymbol{\mu}}_i$ for missing region, to predict the final $N_{comp}$ embeddings $\tilde{\textbf{e}}_i$. Then the output embeddings are sent to MLPs to get latent codes $\tilde{\textbf{z}}_i$ and domain parameters $\tilde{\textbf{A}}_i$ for complete shape representation. Note that the $N_{miss}$ query embeddings $\ddot{\textbf{e}}$ share the same parameter. In addition, center offsets $\tilde{\boldsymbol{\delta}}_i$ are predicted and added to the centers $\tilde{\boldsymbol{\mu}}_i$ for the reason of providing optimal positions for local basis functions. 

To learn the whole local bases, we use four loss functions to supervise the parameters of whole local bases, which are $\mathcal{L}_{sdf}^{dom}$, $\mathcal{L}_{sdf}^{euc}$, $\mathcal{L}_{smooth}$, and $\mathcal{L}_{reg}$. 
$\mathcal{L}_{sdf}^{dom}$ has the same form and function as Eq.~\ref{eq:sdf_loss}. However, only using $\mathcal{L}_{sdf}^{dom}$ will cause problems in learning missing local basis functions. As with prior input embeddings, visible local bases are learned more quickly than the missing ones so that their domains soon expand to a wide range that even covers the missing coordinates, making the weights $\alpha_i$ close to $0$ for missing local basis functions. 
We address the problem by adding the loss function $\mathcal{L}_{sdf}^{euc}$, which simply discards the use of weights. The loss function $\mathcal{L}_{smooth}$ is used to make the transition between two adjacent local bases smooth, and $\mathcal{L}_{reg}$ is used for keeping the coordinate offsets $\tilde{\boldsymbol{\delta}}_i$ small. The final loss functions are shown below,

\begin{equation}
\begin{aligned}
  \mathcal{L}_{inte}
 = \mathcal{L}_{sdf}^{dom} + \mathcal{L}_{sdf}^{euc} + \lambda_1 \mathcal{L}_{smooth} + \lambda_2 \mathcal{L}_{reg},
  \label{eq:all_loss}
\end{aligned}
\end{equation}

\begin{equation}
\begin{aligned}
  \mathcal{L}_{smooth}
 = \frac{1}{|\mathcal{X}|}\sum_{\textbf{x},\textbf{y} \in \mathcal{X},\mathcal{Y}} 
  |f_{p}(\textbf{x})-f_{q}(\textbf{x})|,
  \label{eq:smooth_loss}
\end{aligned}
\end{equation}

\begin{equation}
\begin{aligned}
  \mathcal{L}_{sdf}^{euc}
 = \frac{1}{|\mathcal{X}|}\sum_{\textbf{x},\textbf{y} \in \mathcal{X},\mathcal{Y}} 
  |f_{k}(\textbf{x})-\textbf{y}|, \ 
  \mathcal{L}_{reg}
 = \frac{1}{N_{comp}} \sum_{i \in [N_{comp}]} |\boldsymbol{\delta}_i|,
  \label{eq:sdf_loss_euc}
\end{aligned}
\end{equation}
where $\lambda_1$ is $0.5$, and $\lambda_2$ is $0.01$ in the first epoch of training and $0.0$ in the other epochs. $k$ is defined as the index of the $\boldsymbol{\mu}_i$ which is closest to the query point $\textbf{x}$ in Euclidean distance. $\mathcal{X}$ are the mixture of uniformly sampled points and the points sampled near the complete surface as in DeepSDF~\cite{park2019deepsdf}.

\begin{table*}[t]
  \setlength{\tabcolsep}{2.5mm}
  \caption{Quantitative comparison on 8 trained classes. `Ours (opt.)' means the the optimized results of our method. `PoinTr-NDC' and `SnowflakeNet-NDC' convert the point cloud results of `PoinTr' and `SnowflakeNet' to meshes using Neural Dual Contouring (NDC)~\cite{chen2022neural}. As the reconstructed meshes are not watertight, their IoU scores are not listed.}
  \begin{tabular}{c||l||c c c c c c c c||c}
  \toprule
  Metric & Method & airplane & cabinet & car & chair & lamp & sofa & table & vessel & mean \\
  \midrule
  \multirow{6}{*}{\makecell[c]{IoU $\uparrow$}} & DeepSDF & 0.174 & 0.490 & 0.611 & 0.376 & 0.286 & 0.437 & 0.287 & 0.356 & 0.377 \\
  & IF-Net & 0.811 & 0.656 & 0.845 & 0.765 & 0.676 & 0.825 & 0.646 & 0.799 & 0.753 \\
  & LDIF & 0.637 & 0.657 & 0.785 & 0.630 & 0.454 & 0.760 & 0.534 & 0.639 & 0.637 \\
  & ShapeFormer & 0.662 & 0.586 & 0.802 & 0.629 & 0.526 & 0.747 & 0.552 & 0.671 & 0.647 \\
  & Ours & \textbf{0.853} & \textbf{0.778} & \textbf{0.867} & \textbf{0.830} & \textbf{0.763} & \textbf{0.872} & \textbf{0.766} & \textbf{0.851} & \textbf{0.822} \\
  & Ours (opt.) & 0.843 & \textbf{0.778} & 0.863 & 0.822 & 0.756 & 0.868 & 0.759 & 0.845 & 0.817 \\
  \midrule
  \multirow{8}{*}{\makecell[c]{CD $\downarrow$\\($\times10^{-3}$)}} & DeepSDF & 9.55 & 2.96 & 2.31 & 3.81 & 8.76 & 6.51 & 7.55 & 8.58 & 6.25 \\
  & IF-Net & 0.142 & 1.12 & 1.08 & 0.523 & 0.687 & 0.338 & 1.44 & 0.645 & 0.747 \\
  & LDIF & 0.385 & 1.10 & 0.365 & 0.840 & 2.00 & 0.437 & 1.36 & 0.661 & 0.893 \\
  & ShapeFormer & 0.219 & 2.03 & 0.423 & 0.945 & 0.899 & 0.625 & 1.42 & 0.320 & 0.860 \\
  & PoinTr-NDC & 0.122 & 0.540 & 0.205 & \textbf{0.286} & \textbf{0.315} & 0.252 & 0.615 & 0.146 & \textbf{0.310} \\
  & SnowflakeNet-NDC & 0.128 & \textbf{0.536} & \textbf{0.202} & 0.300 & 0.400 & 0.251 & \textbf{0.580} & 0.173 & 0.321 \\
  & Ours       & 0.0834 & 0.772 & 0.299 & 0.298 & 0.323 & \textbf{0.250} & 0.825 & 0.118 & 0.371 \\
  & Ours (opt.) & \textbf{0.0791} & 0.771 & 0.312 & 0.296 & 0.333 & 0.256 & 0.782 & \textbf{0.116} & 0.368 \\
  \midrule
  \multirow{8}{*}{\makecell[c]{F1 $\uparrow$}} & DeepSDF & 0.238 & 0.375 & 0.402 & 0.346 & 0.262 & 0.326 & 0.320 & 0.308 & 0.322 \\
  & IF-Net & 0.889 & 0.647 & 0.773 & 0.792 & 0.780 & 0.793 & 0.741 & 0.830 & 0.781 \\
  & LDIF & 0.691 & 0.537 & 0.651 & 0.579 & 0.484 & 0.637 & 0.606 & 0.572 & 0.595 \\
  & ShapeFormer & 0.764 & 0.552 & 0.678 & 0.653 & 0.640 & 0.665 & 0.661 & 0.664 & 0.659 \\
  & PoinTr-NDC & 0.826 & 0.642 & 0.737 & 0.728 & 0.699 & 0.767 & 0.725 & 0.788 & 0.739 \\
  & SnowflakeNet-NDC & 0.817 & 0.657 & 0.732 & 0.728 & 0.692 & 0.750 & 0.737 & 0.779 & 0.737 \\
  & Ours & \textbf{0.920} & 0.749 & \textbf{0.795} & \textbf{0.858} & \textbf{0.848} & \textbf{0.848} & 0.841 & \textbf{0.883} & \textbf{0.843} \\
  & Ours (opt.) & 0.918 & \textbf{0.778} & 0.789 & 0.855 & 0.845 & 0.847 & \textbf{0.844} & \textbf{0.883} & 0.841 \\
  \bottomrule
  \end{tabular}
  \label{tab:trained_comparison}
\end{table*}

\subsection{Post Optimization}
\label{sec:post_optimization}

From Sec.~\ref{sec:local_shape_prediction}, we get a collection of local bases defined by latent codes $\textbf{z}_i$, domain parameters $\textbf{A}_i$ and coordinates $\boldsymbol{\mu}_i$ to represent a complete shape. The represented shapes already have good quality (see Tab.~\ref{tab:trained_comparison} and Fig.~\ref{fig:compare}), but they can be further optimized. In short, we optimize $\textbf{z}_i$ and $\boldsymbol{\mu}_i$ to minimize the following loss, where $\lambda_1, \lambda_2, \lambda_3, \lambda_4=1.0, 10.0, 10.0, 0.1$.

\begin{equation}
\begin{aligned}
  \mathcal{L}_{opt}
 = \lambda_1 \mathcal{L}_{face} + \lambda_2 \mathcal{L}_{pos} + \lambda_3 \mathcal{L}_{adj} + \lambda_4 \mathcal{L}_{stable},
  \label{eq:opt_loss}
\end{aligned}
\end{equation}
$\mathcal{L}_{face}$ is used to guarantee the predicted signed-distance values of input points $\mathcal{X}_{in}$ are close to $0$. $\mathcal{L}_{pos}$ is used to ensure the signed distance value of sampled points, whose signed distance value is confirmed as positive, is larger than $0$. $\mathcal{L}_{adj}$ is used to make the signed distance value change smoothly between two adjacent local bases, and $\mathcal{L}_{stable}$ is used to keep the optimized parameters close to the original ones. More details are in the supplementary material.

\begin{table}[t]
  \setlength{\tabcolsep}{4mm}
  \caption{Quantitative comparison on unseen classes. }
  \begin{tabular}{ l || c  c  c }
  \toprule
  Method & IoU$\uparrow$ & \makecell[c]{CD$\downarrow$ ($\times10^{-3}$)} & F1$\uparrow$ \\
  \midrule
  DeepSDF & 0.335 & 6.74 & 0.311 \\
  IF-Net & 0.674 & 0.837 & 0.721 \\
  LDIF & 0.458 & 1.15 & 0.436 \\
  ShapeFormer & 0.572 & 1.02 & 0.607 \\
  PoinTr-NDC & - & \textbf{0.511} & 0.656 \\
  SnowflakeNet-NDC & - & 0.648 & 0.652 \\
  Ours  & \textbf{0.771} & 0.611 & 0.812 \\
  Ours(opt.)  & 0.770 & 0.596 & \textbf{0.818} \\
  \bottomrule
  \end{tabular}
  \label{tab:unseen_comparison}
\end{table}

\section{Experiment}

We execute a series of experiments to evaluate our method for 3D completion. By default, we use $N_{init}=128$, $N_{vis}=64$ and $N_{miss}=32$, and test with unoptimized results. 

\paragraph{Dataset.} The experiments are run on the ShapeNet dataset~\cite{chang2015shapenet}. We preprocess the shapes to make them water-tight following the instructions from Occupancy Networks~\cite{mescheder2019occupancy}. We first generate the $224\times224$ depth scans of 16 random views around the objects. The input point clouds are fetched by reprojecting the 2D pixels and sampling within the 3D points using FPS. We set the number of input points $M=2048$ in all our experiments. Except in the ablation study, we use the official training splits of 8 classes as~\cite{yuan2018pcn} for training, and for testing, we perform two kinds of experiments. One experiment is for the trained 8 classes where we use $3000$ partial inputs randomly sampled from the testing splits of the 8 classes, and the other is for the unknown classes where we use $3000$ partial inputs from 5 unseen classes. As for the ablation study, we use the chair class for training and testing. More details are in the supplementary material.

\paragraph{Metrics.} For the evaluation we use the metrics of Intersection-over-Union (IoU)~\cite{mescheder2019occupancy}, Chamfer L2 Distance (CD)~\cite{mescheder2019occupancy} and F-score\%1 (F1)~\cite{tatarchenko2019single}. If there is no notation, we sampled 100k points on each mesh surface for the computation of CD and F1.

\paragraph{Baselines.}
We compare our method with the state-of-the-art implicit-function based 3D completion methods: IF-Net~\cite{chibane2020implicit}, decoder-only DeepSDF~\cite{park2019deepsdf}, LDIF~\cite{genova2020local}, and ShapeFormer~\cite{yan2022shapeformer}. For a fair comparison, we set the number of shape elements in LDIF to be $96$ which is the same as ours. For the training of ShapeFormer, in which the latent resolution is set to $16^3$, the scaling augmentation is used, or it will fall into overfitting quickly.  We also compare our method with PoinTr~\cite{yu2021pointr} and SnowflakeNet~\cite{xiang2021snowflakenet} which are 3D point cloud completion methods that also use Transformer architectures. All these baseline methods use the same training data and testing data as ours. But the input formats are a little different for some methods. The input of LDIF is a scanned depth map which is the source of our partial point cloud, and the input of IF-Net is a $128\times128\times128$ discrete occupancy grid from the input point cloud.

\subsection{Results on Trained Classes}

As shown in Tab.~\ref{tab:trained_comparison}, our method achieves much better scores than other methods. The IoU and F1 scores of our method outperform the second place by $6.2\%$ and $6.9\%$. We can find that DeepSDF is very difficult to restore the shapes because of the limitation of a single latent code. IF-Net, LDIF, and ShapeFormer achieve better scores but are still not comparable with ours. 
PoinTr-NDC and SnowflakeNet-NDC obtain slightly better scores in CD than ours as the target of PoinTr and SnowflakeNet is to minimize the chamfer distance, but they have lower scores in F1. 
An interesting phenomenon is that after optimization, our method achieves better scores in CD but worse scores in IoU and F1. It illustrates that the initial shapes directly predicted from networks are accurate enough.

\begin{figure*}[t]
  \centering
   \includegraphics[width=1.0\linewidth]{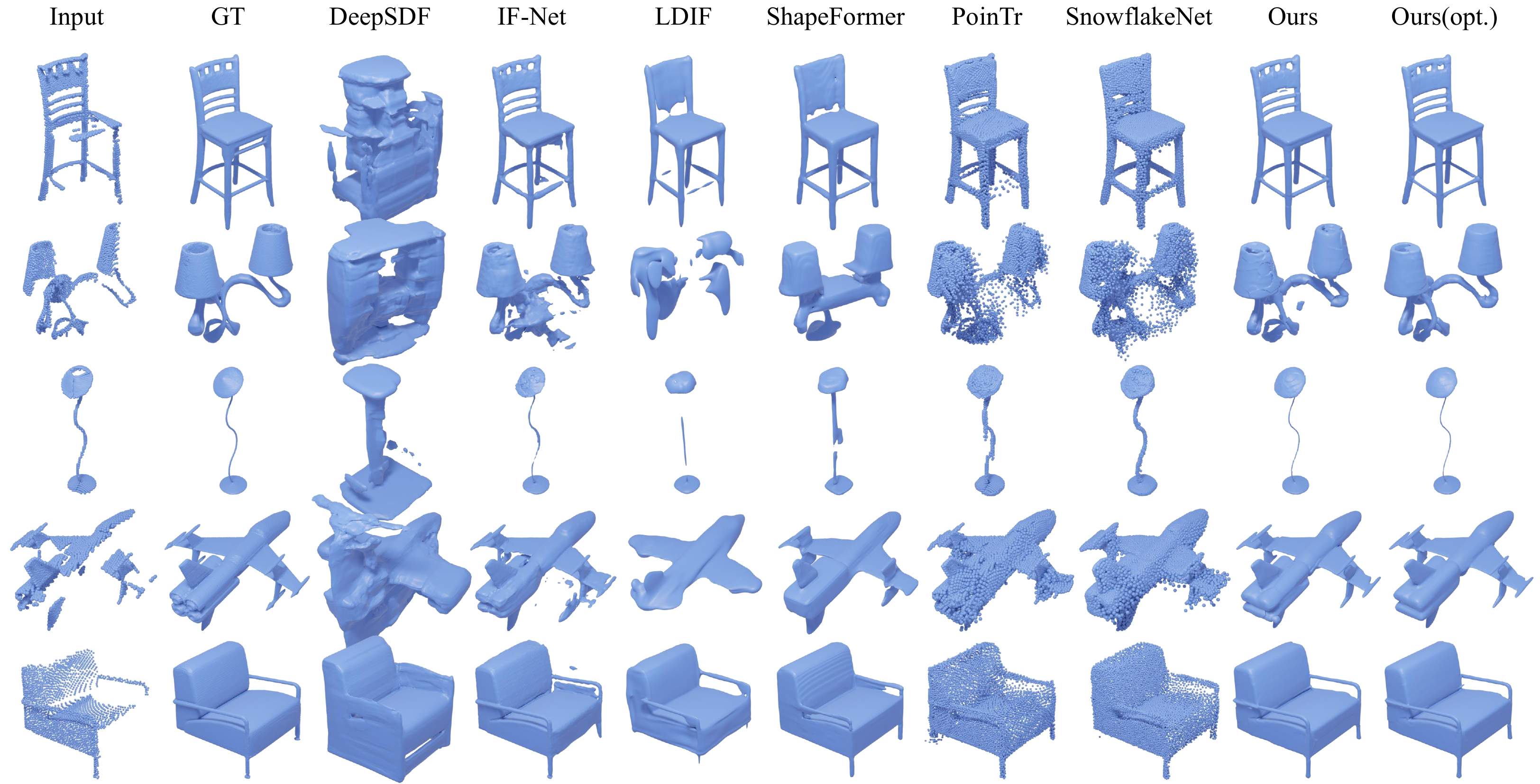}
   \caption{Qualitative comparison with other methods.`Ours (opt.)' means the optimized results of our method.}
   \label{fig:compare}
  \vspace{0.3cm}
\end{figure*}

We also show the qualitative comparison in Fig.~\ref{fig:compare}. 
From the results, we can find that DeepSDF fails to repair shapes when facing complex shapes. 
IF-Net preserves fine details for the visible part because the input points are transferred into a high-resolution voxel grid which saves the detail information in an explicit way, but it cannot avoid the generation of noises for invisible parts.
LDIF, and ShapeFormer fail to recover fine details. We can see that details such as small holes and thin curves are missing in their results. It is mainly because of the low capacity of their shape basis (i.e., Gaussians and vector quantized DIF) for capturing details. 
Obviously, our method owns a better capacity in preserving details for shape completion. For example, in Row 1 and Row 3, our method preserves the correct holes on the chair back and the very thin line correctly. Row 2 shows the power of our method in handling complex surfaces, and it also shows the benefit of post optimization, which is flattening the uneven surface.

\subsection{Results on Unseen Classes}

The results on unseen classes can represent the generalization ability of the shape completing methods on other shapes. From Tab.~\ref{tab:unseen_comparison}, we can find that our method shows outstanding generalization ability in completing the shapes of unseen classes. It is mainly because of our adaptive local bases and the local-to-local translation mechanism which enables predicting missing parts locally. As local parts across different classes may share a similar shape distribution (e.g. the legs of chairs and tables), our method can share the learned local-to-local translation priors from the trained classes with the unseen classes to improve the generalization ability. 
Again, PoinTr-NDC obtains a slightly better score in CD than ours as PoinTr aims to minimize the chamfer distance.
We provide more qualitative comparisons in the supplementary materials.

\begin{table}[t]
  \setlength{\tabcolsep}{0.8mm}
  \caption{Comparison on computational complexity. Floating Point Operations (FLOPs) are counted according to a single forward of each method with one partial input and one query point. }
  \begin{tabular}{c || c  c  c  c  c }
  \toprule
  Methods & DeepSDF & IF-Net & LDIF & ShapeFormer & Ours \\
  \midrule
  Latent Resolution & 1 & $128^3$ & 96 & $16^3$ & 96 \\
  FLOPs & 1.84 M & 20.6 B & 7.07 B & 112.5 B & 1.65 B \\
  \bottomrule
  \end{tabular}
  \label{tab:complexity_comparison}
\end{table}

\begin{table}[t]
  \setlength{\tabcolsep}{2mm}
  \caption{Ablation study on the number of local bases for encoding.}
  \begin{tabular}{c || c  c  c  c}
  \toprule
  Local Basis Number & 128 & 96 & 64 & 32 \\
  \midrule
  IoU $\uparrow$ & \textbf{0.836} & 0.828 & 0.818 & 0.793 \\
  CD ($\times10^{-3}$) $\downarrow$ & \textbf{0.303} & 0.330 & 0.358 & 0.369\\
  F1 $\uparrow$ & \textbf{0.866} & 0.861 & 0.852 & 0.829\\
  \bottomrule
  \end{tabular}
  \label{tab:aldif_number}
\end{table}

\subsection{Comparison with PatchRD}

\begin{figure*}[t]
  \centering
   \includegraphics[width=1.0\linewidth]{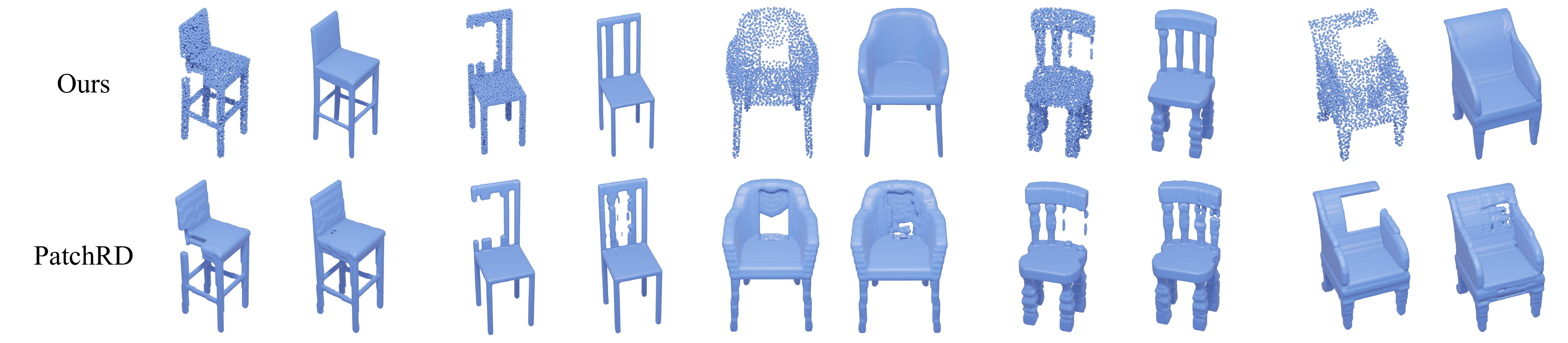}
   \caption{Qualitative comparison with PatchRD. For each pair, the left is the input and the right is the output. }
   \label{fig:comp_patchRD}
\end{figure*}

PatchRD~\cite{sun2022patchrd} is a voxel-based completion method. While it uses 3D convolutional encoding, its output is based on retrieval rather than direct convolutional decoding. We follow PatchRD to crop out small regions to generate input data, and perform qualitative comparison with it. The results are shown in Fig.~\ref{fig:comp_patchRD}. Our method can produce more accurate and smoother results than PatchRD.

\subsection{Different Levels of Completeness}

\begin{figure*}[t]
  \centering
   \includegraphics[width=1.0\linewidth]{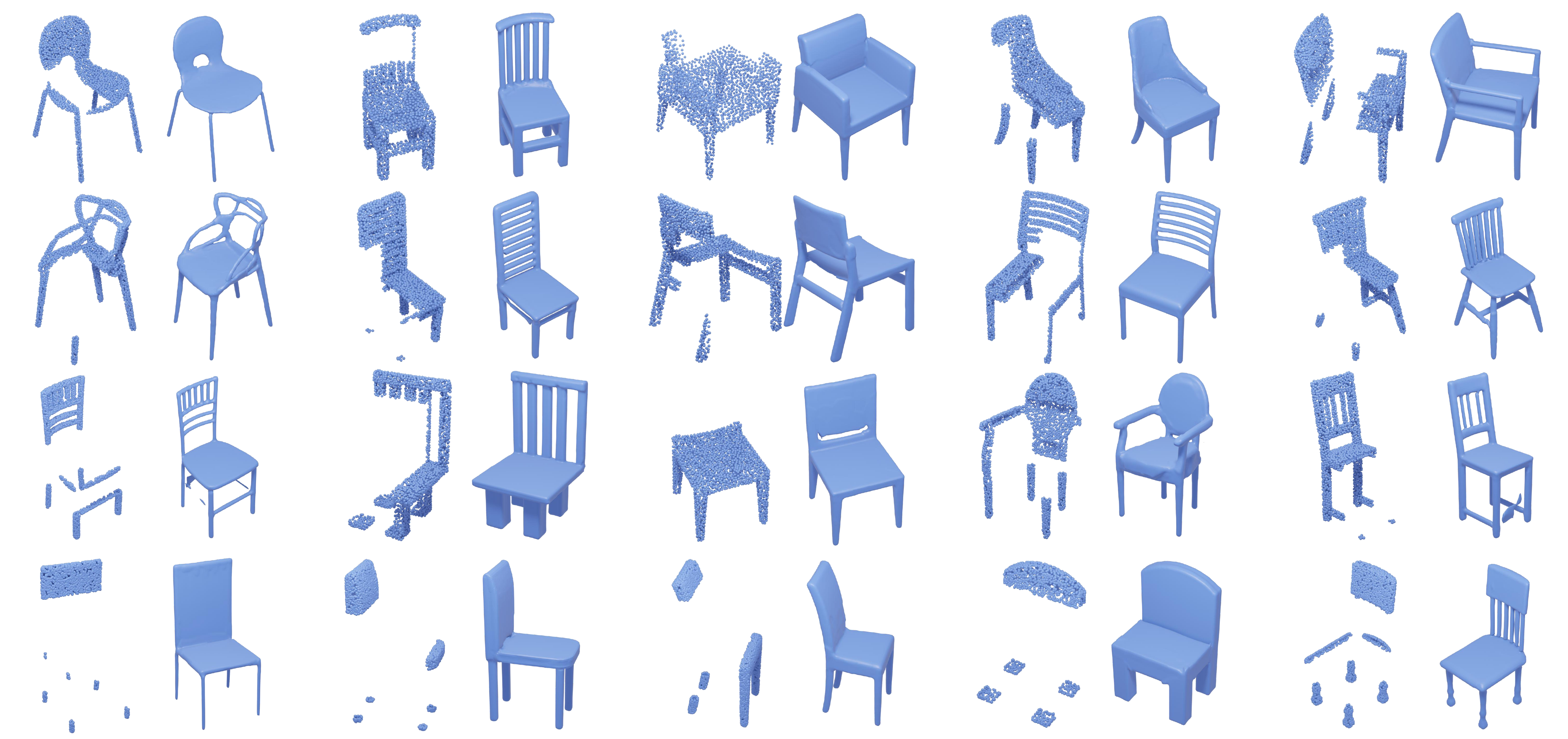}
   \caption{Qualitative results on point clouds with different levels of completeness. For each pair, the left is the input and the right is the output.} 
   \label{fig:crop_res}
  \vspace{0.3cm} 
\end{figure*}

In order to validate the robustness of our method, we conduct experiments on input points with different levels of completeness, as shown in Fig.~\ref{fig:crop_res}. Our method is robust to different levels of completeness and can preserve fine details in general. But we still observe floating crossbars near the chair legs (Column 1\&5 in Row 3). It may be caused by the ambiguity of crossbar existence in chairs. One interesting future work is how to avoid such ambiguities.

\begin{table}[t]
  \setlength{\tabcolsep}{3mm}
  \caption{Ablation study on smooth loss and domain-based downsampling. }
  \begin{tabular}{ c c || c  c  c }
  \toprule
  \makecell[c]{Smooth\\Loss} & \makecell[c]{Domain-based\\Downsampling} & IoU$\uparrow$ & \makecell[c]{CD $\downarrow$\\($\times10^{-3}$)} & F1$\uparrow$ \\
  \midrule
  \ding{56} & \ding{52} & 0.797 & 0.409 & 0.832 \\
  \ding{52} & \ding{56} & \textbf{0.808} & 0.394 & 0.841 \\
  \ding{52} & \ding{52} & \textbf{0.808} & \textbf{0.364} & \textbf{0.844} \\
  \bottomrule
  \end{tabular}
  \label{tab:loss_ablation}
\end{table}

\subsection{Results on Real Scans}

\begin{figure*}[t]
  \centering
   \includegraphics[width=1.0\linewidth]{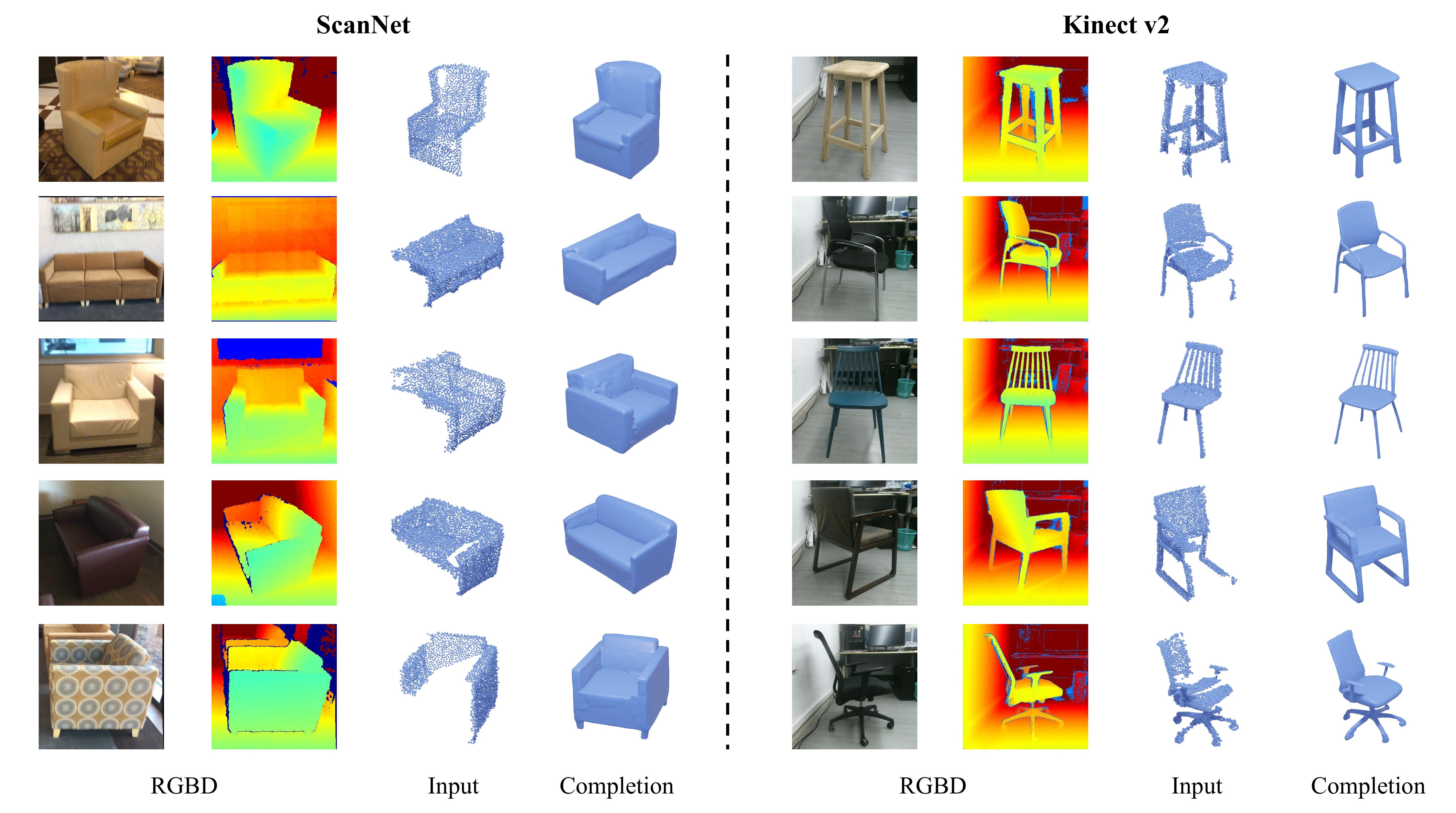}
   \caption{Shape completion results on real-world scans.}
   \label{fig:real_comp}
\end{figure*}

We have investigated the completion ability of our method on real-world scans. Fig.~\ref{fig:real_comp} shows the results of different objects from ScanNet~\cite{dai2017scannet} and scanned by ourselves using Kinect v2. Our method achieves high-quality completion results on real scan data.

\subsection{Comparison on Computational Cost}

Tab.~\ref{tab:complexity_comparison} gives the comparison of computational cost among DIF-based baseline methods and ours. Although DeepSDF achieves the fewest latent codes and FLOPs, it fails in completing complex shapes as illustrated above. Among the other methods, it can be easily found that our method has a much lower computational cost, because we use a compact shape representation which requires a small number of latent codes and avoids the use of 2D and 3D convolutions which consume a lot of computations in LDIF, IF-Net, and ShapeFormer.

\subsection{Ablation Studies} 

\begin{figure*}[t]
  \centering
   \includegraphics[width=1.0\linewidth]{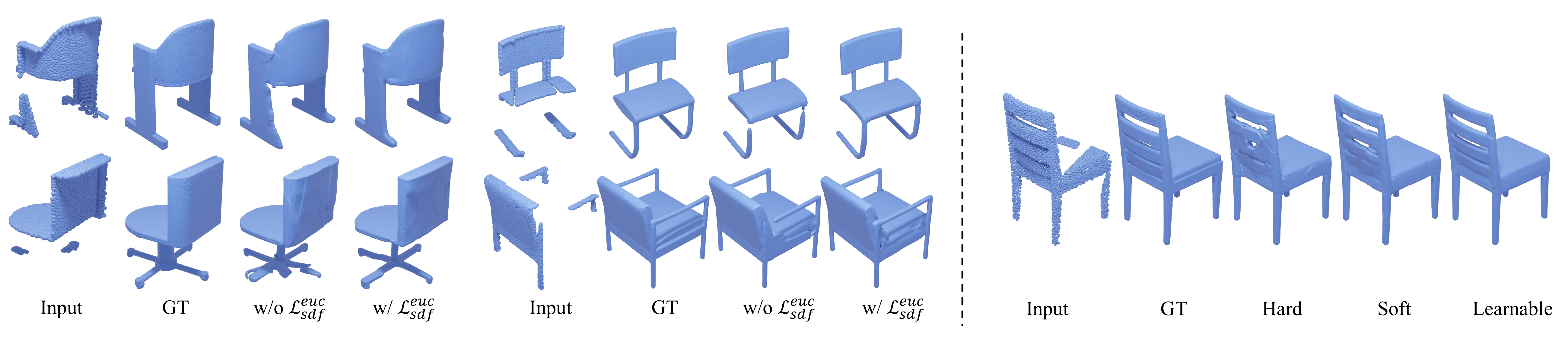}
   \caption{Ablation study on $\mathcal{L}_{sdf}^{euc}$ (left) and different weight strategies (right). }
   \label{fig:ablation}
\end{figure*}

\paragraph{Weight Strategies.}
To prove that our learnable weights $\alpha_i$ can improve the learning of DIF-based basis functions and therefore help with preserving details, we compare the completion results with different weight strategies. The comparison results can be found in Fig.~\ref{fig:ablation} (right). The learnable strategy is what we used, while the soft strategy means using the same formulation as Eq.~\ref{eq:sdf_form_2} but the weights $\alpha_i$ are unlearnable which means the domain parameters $\text{A}_i$ and the offsets $\boldsymbol{\delta}_i$ are fixed (we set $\textbf{R}_i$ to be an identity matrix, $\boldsymbol{\sigma}_i=[500, 500, 500]$ and $\boldsymbol{\delta}_i$=0.). The hard strategy is more direct that $\alpha_i$ is $1$ if the center $\boldsymbol{\mu}_i$ is closest to the query point $\textbf{x}$ or $0$ otherwise. Obviously, the learnable strategy can help provide more details.

\paragraph{Local Basis Number.}
In order to verify that our local bases are compact, we do an ablation study using different numbers of local bases during partial point cloud encoding. Specifically, we fix the initial local basis number ($N_{init}$) to $128$ but vary the downsampling number ($N_{vis}$) for partial points encoding and shape completion. From Tab.~\ref{tab:aldif_number}, we can find that the scores keep close with different local basis numbers. When the local basis number decreases from $128$ to $32$, the scores only drop a little. That means our local bases can be learned adaptively to faithfully represent the shapes.

\paragraph{Nearest Basis Number.}
In our framework, the signed distance of a query point is determined by two nearest local basis functions. We have also tried to use three nearest bases, and the IoU score decreases by $0.2\%$ and F1 score remains the same. Therefore, two nearest local bases are sufficient to express the signed distance.

\paragraph{Loss Functions.}
We conduct ablation studies on different loss functions. As shown in Row 1 and 3 in Tab.~\ref{tab:loss_ablation}, by using the smooth loss, the IoU score increases by $1.1\%$ and F1 score increases by $1.2\%$. As shown in Fig~\ref{fig:ablation} (left), without $\mathcal{L}_{sdf}^{euc}$, the prediction of missing parts degrades. Removing $\mathcal{L}_{reg}$ makes the training unable to converge. 

\paragraph{Domain-based Downsampling.}
We validate the domain-based downsampling by replacing it with uniform downsampling. As shown in Row 2 and 3 in Tab.~\ref{tab:loss_ablation}, 
domain-based downsampling obtains the best scores.

\section{Conclusion and Limitation}

We have proposed a new shape completion method based on implicit function with adaptive local basis functions. These local basis functions provide an effective and efficient compact representation for complex shapes, preserving geometric details while reducing computational costs. One limitation of our method is that the whole local bases prediction does not guarantee to recover the target topology of the shape, shown in (Fig.~\ref{fig:overall_pipeline} (d)) where the cross-bar is not tightly connected with one leg of the chair. This happens when the connectivity of the local bases in the missing region is different from the ground-truth. In the future, we will look into augmenting our method with a high-level graph-based representation that focuses on the global topology of the shape.

\begin{acks}
The authors would like to thank reviewers for their insightful comments. This work was supported by NSF China (62227806), the XPLORER PRIZE, and the 100 Talents Program of Zhejiang University.
\end{acks}

\bibliographystyle{ACM-Reference-Format}
\bibliography{sample-bibliography}

\clearpage

\clearpage
\clearpage

\appendix

\section{Implementation Details}

\subsection{Networks Details}

In Tab.~\ref{tab:detail_architecture}, we show the detailed network architecture of the PointNet++ encoder~\cite{qi2017pointnet++} and head networks. For the PointNet++ encoder, the sampling layer performs farthest point sampling (FPS) to the input points to choose a subset of input points, and the grouping layer uses $K$ nearest neighbors (KNN) to find a fixed number of neighboring points. The hierarchical sampling and grouping operations in the PointNet++ encoder guarantee that the output key point embeddings focus on the local region features. 

\begin{table*}[h]
  \setlength{\tabcolsep}{2mm}
  \caption{The detailed architecture information of our method. $N$ is the number of local bases. For linear layer (LinearLayer and LinearLayer*), `i' and `o' stand for input channel size and output channel size. For convolutional layer (ConvLayer), the `k', `s', and `p' stand for kernel size, stride, and padding. For grouping layer (Grouping), `n' stands for the number of points in the neighborhood of centroid points, and the input consists of two point sets, which are the point set with features before sampling (left) and the coordinates of a set of centroids (right). Also `LinearLayer' denotes a single linear layer, while `LinearLayer*' denotes the composition of linear layer + layer normalization + leaky ReLU. `ConvLayer' denotes the composition of convolutional layer + group normalization + leaky ReLU. \ }
  \begin{tabular}{p{4.5cm} p{2.0cm} p{4.0cm} p{4.0cm}}
  \toprule
   Layer Name & Notes & Input Size & Output Size \\
  \midrule
  \midrule
  \textbf{PointNet++ Encoder} & & & \\
    LinearLayer & i3o16  & $2048\times3$            & $2048\times16$ \\
    Grouping    & n16 & $2048\times(3+16)/2048\times3$ & $2048\times16\times19$ \\
    ConvLayer   & k1s0p0 & $2048\times16\times19$   & $2048\times16\times32$ \\
    ConvLayer   & k1s0p0 & $2048\times16\times32$   & $2048\times16\times64$ \\
    ConvLayer   & k1s0p0 & $2048\times16\times64$   & $2048\times16\times128$ \\
    MaxPooling  &        & $2048\times16\times128$  & $2048\times128$ \\
    Sampling    &        & $2048\times3$            & $512\times3$  \\
    Grouping    & n16 & $2048\times(3+128)/512\times3$ & $512\times16\times131$ \\
    ConvLayer   & k1s0p0 & $512\times16\times131$   & $512\times16\times128$ \\
    ConvLayer   & k1s0p0 & $512\times16\times128$   & $512\times16\times128$ \\
    ConvLayer   & k1s0p0 & $512\times16\times128$   & $512\times16\times128$ \\
    MaxPooling  &        & $512\times16\times128$   & $512\times128$ \\
    Sampling    &        & $512\times3$             & $128\times3$  \\
    Grouping    & n16 & $512\times(3+128)/128\times3$  & $128\times16\times131$ \\
    ConvLayer   & k1s0p0 & $128\times16\times131$   & $128\times16\times128$ \\
    ConvLayer   & k1s0p0 & $128\times16\times128$   & $128\times16\times128$ \\
    ConvLayer   & k1s0p0 & $128\times16\times128$   & $128\times16\times128$ \\
    MaxPooling  &        & $128\times16\times128$   & $128\times128$ \\
    LinearLayer & i128o256 & $2048\times256$        & $128\times256$ \\
  \midrule
  \midrule
  \textbf{Latent Code Head} & & & \\
  LinearLayer*  & i256o256  & $N\times256$          & $N\times256$ \\
  LinearLayer*  & i256o256  & $N\times256$          & $N\times256$ \\
  LinearLayer   & i256o256  & $N\times256$          & $N\times256$ \\
  \midrule
  \midrule
  \textbf{Scaling Head} & & & \\
  LinearLayer*  & i256o256  & $N\times256$          & $N\times256$ \\
  LinearLayer*  & i256o256  & $N\times256$          & $N\times256$ \\
  LinearLayer   & i256o3    & $N\times256$          & $N\times3$ \\
  \midrule
  \midrule
  \textbf{Rotation Head} & & & \\
  LinearLayer*  & i256o256  & $N\times256$          & $N\times256$ \\
  LinearLayer*  & i256o256  & $N\times256$          & $N\times256$ \\
  LinearLayer   & i256o6    & $N\times256$          & $N\times6$ \\
  \midrule
  \midrule
  \textbf{Offset Head} & & & \\
  LinearLayer*  & i256o256  & $N\times256$          & $N\times256$ \\
  LinearLayer*  & i256o256  & $N\times256$          & $N\times256$ \\
  LinearLayer   & i256o3    & $N\times256$          & $N\times3$ \\
  \midrule
  \midrule
  \textbf{Coordinates Head} & & & \\
  LinearLayer*  & i256o512  & $256$                 & $512$ \\
  LinearLayer*  & i256o512  & $512$                 & $512$ \\
  LinearLayer   & i512o96   & $512$                 & $N_{miss}\times3$ \\
  \bottomrule
  \end{tabular}
  \vspace{0.3cm}
  \label{tab:detail_architecture}
\end{table*}

Given $N$ input embeddings with feature size $256$, four head networks are used to decode them into the required parameters for local bases, which are latent codes $\textbf{z}_i$, domain parameters $\textbf{A}_i$, and offsets $\boldsymbol{\delta}_i$. Note that the domain parameter $\textbf{A}_i$ consists of sigma $\boldsymbol{\sigma}_i$ and rotation $\textbf{R}_i$, so there are two head networks to predict these two parameters separately. Also, the head networks used in the partial point cloud encoding and the whole local bases prediction share the same architecture but have different weights. In addition to these four head networks, the coordinate head network takes as input one $256$-dimension embedding and predicts $N_{miss}\times3$ dimension features that can be reshaped as $N_{miss}$ coordinates for the missing region.

For the implicit decoder, we follow the same architecture as DeepSDF~\cite{park2019deepsdf}. Both Missing Centers Transformer and Local Bases Transformer consist of several self-attention blocks as the encoder in~\cite{vaswani2017attention}. We use $3$ and $6$ blocks for Missing Centers Transformer and Local Bases Transformer, respectively. All of these blocks have $8$ heads self-attention, and the embedding dimension is 256. All input coordinates $\boldsymbol{\mu}_i$ are mapped into $256$-dimension positional embeddings by $2$ linear layers.

\subsection{Training Details}

Directly training the whole model makes it hard to converge, so we adopt a two-phase training. The first phase is to train the networks and parameters used in partial point cloud encoding and missing centers prediction simultaneously. The used loss function is $\mathcal{L}_{sdf} + \lambda\mathcal{L}_{cham}$ with $\lambda=0.01$. It is worth mentioning that during training, the embeddings $\textbf{e}_i$ and coordinates $\boldsymbol{\mu}_i$ after downsampling will be inputted into Missing Centers Transformer in forward propagation, but their gradients will not be returned from the Transformer in backward propagation in order to avoid the influence of missing centers prediction on partial point cloud encoding. After the first phase of training, those trained networks and parameters are fixed, and the next phase is to train the query embedding $\ddot{\textbf{e}}$ and Local Bases Transformer, as well as the head networks and implicit decoder used in whole local bases prediction, with the loss function $\mathcal{L}_{inte}$. Before the start of the second phase training, the weight of the untrained implicit decoder and head networks can be initialized with the ones trained in the first phase who share the same architectures.

In the both training phases, we utilize Adam optimizer and multi-step learning rate scheduler to train the networks and parameters with the batch size $16$ and the initial learning rate $0.0005$. In the main experiments, during each phase we perform training for $20$ epochs and reduce the learning rate  by $50\%$ in the epochs of $\{10, 15, 18\}$. While in the ablation study, during each phase we perform training for $30$ epochs and reduce the learning rate by $50\%$ in the epochs of $\{15, 22, 27\}$. 
More implementation details can be found in the supplementary materials.

In the experiments, the trained classes for ShapeNet~\cite{chang2015shapenet} include airplane, cabinet, car, chair, lamp, sofa, table, and vessel, while the unseen classes include telephone, loudspeaker, display, bench, and rifle.

\subsection{Post Optimization}

During inference, after predicting the parameters of local bases from networks, we can further optimize $\textbf{z}_i$ and $\boldsymbol{\mu}_i$ to get a more accurate and smoother shape by minimizing the following loss. We utilize Adam optimizer to perform optimization for $1000$ iterations with a learning rate of $0.001$.

\begin{equation}
\begin{aligned}
  \mathcal{L}_{opt}
 = \lambda_1 \mathcal{L}_{face} + \lambda_2 \mathcal{L}_{pos} + \lambda_3 \mathcal{L}_{adj} + \lambda_4 \mathcal{L}_{stable}.
  \label{eq:opt_loss_2}
\end{aligned}
\end{equation}

The function of $\mathcal{L}_{face}$ shown below is to guarantee the predicted signed-distance values of input points $\mathcal{X}_{in}$ are close to $0$. We use a margin $\epsilon$ to relax the loss that only the items with $|sdf(\textbf{x})|$ larger than $\epsilon$ will be punished. 

\begin{equation}
\begin{aligned}
  \mathcal{L}_{face}
 = \frac{1}{|\mathcal{X}_{in}|} \sum_{\textbf{x} \in \mathcal{X}_{in}} 
    min(|sdf(\textbf{x})|-\epsilon, 0.0)^2.
  \label{eq:face_loss}
\end{aligned}
\end{equation}

The function of $\mathcal{L}_{pos}$ shown below is to ensure the signed distance value of positive points $\mathcal{X}_{pos}$ is larger than $0$. Similar to Eq.~\ref{eq:face_loss}, a margin $\epsilon$ is also used to relax the loss. The positive points $\mathcal{X}_{pos}$ are the points that are sampled around the key point coordinates and whose signed distance value is confirmed as positive. For example, if the sampled points are located between the camera and the observed depth map or outside silhouette of the depth map, they will be regarded as positive.

\begin{equation}
\begin{aligned}
  \mathcal{L}_{pos}
 = \frac{1}{|\mathcal{X}_{pos}|} \sum_{\textbf{x} \in \mathcal{X}_{pos}} 
    min(-sdf(\textbf{x})-\epsilon, 0.0)^2.
  \label{eq:pos_loss}
\end{aligned}
\end{equation}

The function of $\mathcal{L}_{adj}$ shown below is to make the signed distance value change smoothly between two adjacent local bases. In the function, $\mathcal{X}_{samp}$ are the points sampled around the key point coordinates. $p$ and $q$ are the same as the ones introduced in Sec. 3 of the main paper. $\omega_1(\textbf{x})$ and $\omega_2(\textbf{x})$ are the weight factors, that $\omega_1(\textbf{x})$ gets larger when $\textbf{x}$ is close to the surface and $\omega_2(\textbf{x})$ gets larger when $g_p(\textbf{x})$ and $g_q(\textbf{x})$ are close. In practice, we set $a_1=10000$ and $a_2=1000$.

\begin{equation}
\begin{aligned}
\label{eq:adj_loss}
  \mathcal{L}_{adj}
 =& \frac{1}{|\mathcal{X}_{samp}|} \sum_{\textbf{x} \in \mathcal{X}_{samp}} 
    \omega_1(\textbf{x})\omega_2(\textbf{x})(f_{\phi,p}(\textbf{x})-f_{\phi,q}(\textbf{x}))^2, \\
  \text{where }
    &\omega_{1}(\textbf{x}) = 
     \exp(-a_1 \min(f_{\phi,p}(\textbf{x}), f_{\phi,q}(\textbf{x}))^2) \\
    &\omega_{2}(\textbf{x}) = 
     \exp(-a_2 (g_p(\textbf{x})-g_q(\textbf{x}))^2).
\end{aligned}
\end{equation}

The last loss $\mathcal{L}_{stable}$ shown below is to keep the optimized parameters close to the original latent codes $\hat{\boldsymbol{\mu}}_i$ and coordinates $\hat{\textbf{z}}_i$.

\begin{equation}
\begin{aligned}
  \mathcal{L}_{stable}
 = \frac{1}{N_{comp}} \sum_{i \in [N_{comp}]}
 ||\boldsymbol{\mu}_i - \hat{\boldsymbol{\mu}}_i||^2
 +||\textbf{z}_i - \hat{\textbf{z}}_i||^2
  \label{eq:const_loss}.
\end{aligned}
\end{equation}

\section{Experiments}

\begin{table}[t]
  \setlength{\tabcolsep}{3mm}
  \caption{Comparison with PoinTr and SnowflakeNet.} 
  \begin{tabular}{ l || c c | c c }
  \toprule
   & \multicolumn{2}{c|}{Trained} &  \multicolumn{2}{c}{Unseen} \\
  \cline{2-5}
  & \makecell[c]{CD $\downarrow$\\($\times10^{-3}$)} & F1 $\uparrow$ & \makecell[c]{CD $\downarrow$\\($\times10^{-3}$)} & F1 $\uparrow$ \\
  \midrule
  PoinTr & 0.279 & 0.585 & 0.405 & 0.558 \\
  SnowflakeNet & \textbf{0.252} & 0.628 & \textbf{0.374} & 0.633 \\
  Ours & 0.447 & \textbf{0.637} & 0.678 & \textbf{0.640} \\
  \bottomrule
  \end{tabular}
  \label{tab:pointr_comparison}
\end{table}

\subsection{Comparison with PoinTr and SnowflakeNet}

We also compare our method directly with the point-based completion method PoinTr and SnowflakeNet~\cite{xiang2021snowflakenet}  which also adopts a Transformer architecture. As the number of output points is fixed to be $8192$ in PoinTr and SnowflakeNet, we sample the same number of points on the meshes of our method for computing CD and F1. The results are listed in Tab.~\ref{tab:pointr_comparison}. We can find that the PoinTr and SnowflakeNet have better scores in CD, and our method achieves better scores in F1. That means the whole point cloud generated by PoinTr and SnowflakeNet obtain lower average errors in L2 distance, but there are numbers of outliers that reduce the F1 score. Our methods can produce more stable results with fewer outliers. It can also be proven in the visualization results shown in the main paper.
Notably, for both methods, there are small gaps between the scores on trained classes and unseen classes. It further proves that the generalization ability of shape completion can be improved by building local-to-local translation modules among local shape representations. 

\begin{table}[t]
  \setlength{\tabcolsep}{4mm}
  \caption{Shape reconstruction results. `Lat. Res.' means latent resolution, and `ShapeF.' means ShapeFormer.} 
  \begin{tabular}{ l || c || c  c  c }
  \toprule
  Method & \makecell[c]{Lat.\\Res.} & IoU$\uparrow$ & \makecell[c]{CD $\downarrow$\\($\times10^{-4}$)} & F1$\uparrow$ \\
  \midrule
  DeepSDF& $1$ & 0.560 & 18.5 & 0.425 \\
  IFNet & $128^3$ & 0.915 & \underline{0.312} & \underline{0.985} \\
  LDIF & $32$ & 0.769 & 3.25 & 0.834 \\
  ShapeF. & $16^3$ & 0.814 & 0.641 & 0.884 \\
  ShapeF. & $32^3$ & 0.844 & 0.671 & 0.928 \\
  3DILG & $128$ & 0.864 & 0.782 & 0.920 \\
  3DILG & $256$ & 0.895 & 0.485 & 0.956 \\
  3DILG & $512$ & 0.920 & 0.374 & 0.978 \\
  Ours & $32$ & 0.892 & 0.804 & 0.936 \\
  Ours & $64$ & 0.928 & 0.437 & 0.968 \\
  Ours & $128$ & \underline{0.948} & 0.348 & 0.981 \\
  Ours & $256$ & \textbf{0.957} & \textbf{0.274} & \textbf{0.987} \\
  \bottomrule
  \end{tabular}
  \label{tab:mesh_reconstruction}
\end{table}

\subsection{Shape Reconstruction on Complete Input Points}

To show the efficiency of our shape representation method, we perform shape reconstruction using complete point clouds and compared the results with other DIF-based methods. For this task, we only use the `Visible Points Encoding' module and use the loss function $\mathcal{L}_{sdf} + \lambda\mathcal{L}_{smooth}$ in which $\lambda=0.5$. Here, our post optimization is not used. All the experiments are run on the chair class of ShapeNet dataset~\cite{chang2015shapenet}, and the results are shown in Tab.~\ref{tab:mesh_reconstruction}. The methods for comparison include DeepSDF~\cite{park2019deepsdf}, IF-Net~\cite{chibane2020implicit}, LDIF~\cite{genova2020local}, ShapeFormer~\cite{yan2022shapeformer}, and 3DILG~\cite{zhang20223dilg}, which also adopt deep implicit functions and auto-encoder architectures. Our method achieves the best scores when using $256$ local bases. Notably, our method can achieve similar scores with much fewer bases. For example, our method with $128$ local bases can achieve similar scores as IFNet with $128^3$ resolution, and our method with $64$ local bases can achieve similar scores as 3DILG with $512$ latent features. Even with $32$ local bases, our method outperforms LDIF and ShapeFormer in most metrics.

\subsection{More Qualitative Results}

\begin{figure*}[t]
  \centering
   \includegraphics[width=0.9\linewidth]{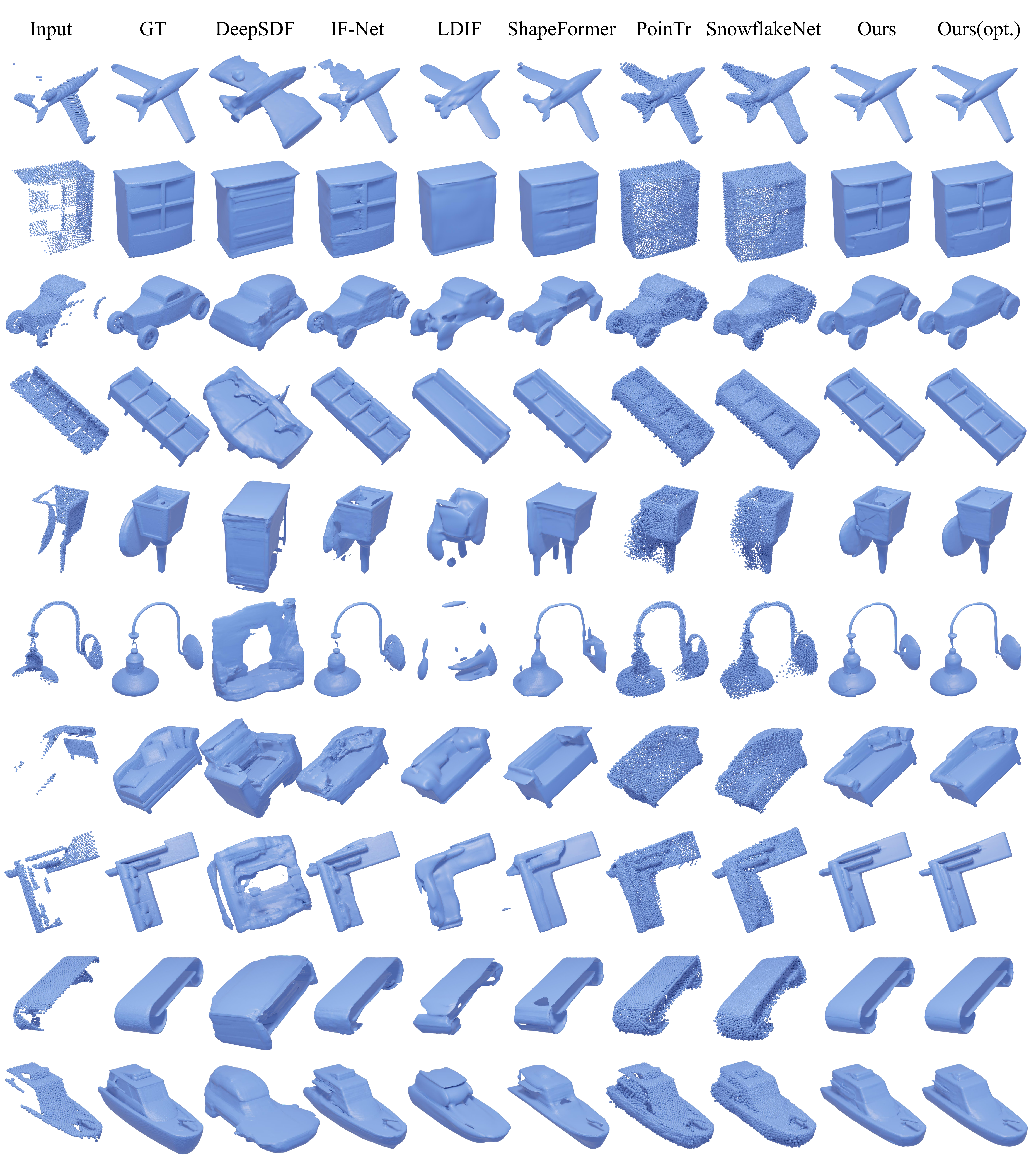}
   \caption{More qualitative comparison with other methods on trained classes.`Ours (opt.)' means the optimized results of our method.}
   \label{fig:trained_comp_supp}
\end{figure*}

\begin{figure*}[t]
  \centering
   \includegraphics[width=0.9\linewidth]{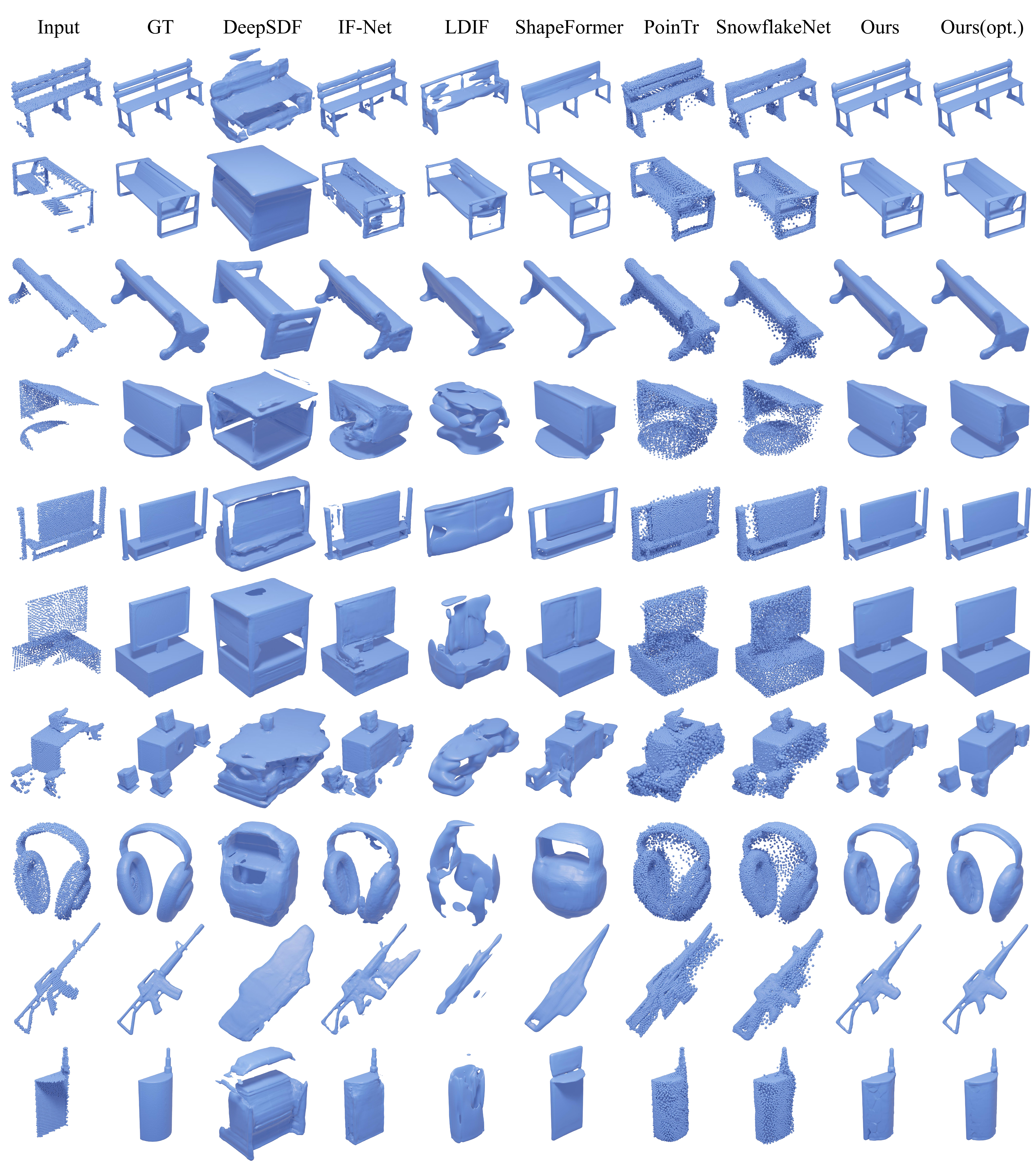}
   \caption{More qualitative comparison with other methods on unseen classes.`Ours (opt.)' means the optimized results of our method.}
   \label{fig:unseen_comp_supp}
\end{figure*}

We show more qualitative comparisons on the trained and unseen classes in Fig.~\ref{fig:trained_comp_supp} and Fig.~\ref{fig:unseen_comp_supp}. We can find that our completion results clearly outperform the ones from other methods. DeepSDF~\cite{park2019deepsdf} fails to generate reasonable shapes when meeting complex cases. LDIF~\cite{genova2020local} and ShapeFormer~\cite{yan2022shapeformer} are difficult to preserve details. Also, ShapeFormer may generate shapes that do not match the input points. PoinTr~\cite{yu2021pointr} may generate outlier points. IF-Net~\cite{chibane2020implicit} can preserve fine details for visible parts but is difficult to predict smooth results for invisible parts. In contrast, our method can predict reasonable shapes for invisible parts while preserving fine details. Fig.~\ref{fig:unseen_comp_supp} demonstrates the strong generalization ability of our method.


\end{document}